\documentclass[review]{elsarticle}
\usepackage{upgreek}
\usepackage{multirow}
\usepackage{hyperref}
\usepackage[table]{xcolor}
\usepackage{longtable}
\usepackage{array}
\usepackage{caption}
\usepackage{graphicx}
\graphicspath{ {./images/} }

\journal{ArXiv}

\biboptions{authoryear}

\begin{document}

\begin{frontmatter}

\title{State of the art applications of deep learning within tracking and detecting marine debris: A survey.}

\author{Zoe Moorton\fnref{myfootnote}\corref{mycorrespondingauthor}}
\author{Dr Zeyneb Kurt\fnref{myfootnote}}
\author{Dr Wai Lok Woo\fnref{myfootnote}}
\address{Department of Computer and Information Sciences,
University of Northumbria, Newcastle Upon Tyne, UK}


\author[mymainaddress]{zoemoorton@gmail.com\corref{mycorrespondingauthor}}
\cortext[mycorrespondingauthor]{Corresponding author}

\begin{abstract}
Deep learning techniques have been explored within the marine litter problem for approximately 20 years but the majority of the research has developed rapidly in the last five years. We provide an in-depth, up to date, summary and analysis of 28 of the most recent and significant contributions of deep learning in marine debris.\\
From cross referencing the research paper results, the YOLO family significantly outperforms all other methods of object detection but there are many respected contributions to this field that have categorically agreed that a comprehensive database of underwater debris is not currently available for machine learning. Using a small dataset curated and labelled by us, we tested YOLOv5 on a binary classification task and found the accuracy was low and the rate of false positives was high; highlighting the importance of a comprehensive database.\\
We conclude this survey with over 40 future research recommendations and open challenges. 
\end{abstract}

\begin{keyword}
\texttt marine debris, artificial intelligence, object detection, remote sensing, data acquisition
\end{keyword}

\end{frontmatter}


\section{Introduction}
The most represented method of approaching the marine debris problem so far has been object detection for floating debris and tracking it with remote sensing.\\
Within this paper; the authors use various terms such as synthetic debris, anthropogenic debris litter, rubbish or trash of which are all intended to have the same definition. Quite simply, these items are disposed of by humans and have found their way into most water systems. A challenging problem as \cite{Watanabe2019} explains that marine environments have more complex and variable backgrounds compared to land.\\
Although, we mean to specifically address marine debris, this research also explores some techniques that have been used in other water bodies such as rivers. It is also worth noting that this survey only covers debris of a macro size and larger. This means that anything less than 5mm has not been included.\
\subsection{Background}
Debris in marine environments has exponentially increased which continues to have destructive results on marine life, the environment and even human health. There is an overwhelming plethora of literature that documents the direct impact of synthetic debris on the marine ecosystem.\\ 
Inhabitants of marine settings are frequently found entangled within nets and other materials; or have been discovered mistaking debris as prey or an object to play with, consequently ingesting the toxic materials \citep{McAdam2017}. In fact, a quantitative overview of marine fauna ingesting waste by \cite{Kühn2020} has found that over 700 species are confirmed to eat plastic. Additionally, a review on ghost gear by \cite{Stelfox2016} confirmed that up until the time of publication, over 40 species had been recorded entangled in ghost fishing gear alone. Ghost fishing gear is defined by \cite{NOAA2023a} as “any discarded, lost, or abandoned, fishing gear in the marine environment. This gear continues to fish and trap animals, entangle and potentially kill marine life, smother habitat, and act as a hazard to navigation.”  Unfortunately, due to the consequence of trapped animals either rapidly sinking or being consumed by predators, we are unable to detect or fairly estimate the number of entangled animals \citep{Sharma2017}, \citep{Allen2012}, \citep{Laist1997},  \citep{Quayle1992}, therefore, the data reported so far can only be a percentage of the scale.\\
Often mistaken for prey or consumed within foraging habits \citep{NOAA2023b}, it is not difficult to see why so many creatures confuse debris as food; some studies have even observed that due to their size; microplastics resemble phytoplankton \citep{Boerger2010} and due to their texture; plastic bags are confused for jellyfish by turtles \citep{Bugoni2001}, \citep{Tomás2002}.\\
Ingesting marine debris results in lethal consequences \citep{Wright2013}, \citep{Pawar2016} as plastic fragments absorbs toxic materials and are highly contaminated with PCB (biphenyl polychlorinated), polypropylene, polyethylene, polystyrene, heavy metals and other noxious substances \citep{Mato2001}. These substances are leached into the consumer \citep{Talsness2009} poisoning the marine fauna, it will cause alternative repercussions including reproductive disorders, hormone changes, higher disease risk or most commonly; obstructing the gastrointestinal tract, resulting in starvation and fatality.\citep{Ryan1988}, \citep{Lee2001}, \citep{Oberdörster2001}, \citep{Derraik2002}, \citep{Rahman2004}, \citep{Plot2011}.\\
Ultimately, plastic debris ends up sinking to the seafloor, where it may stay for hundreds of years \citep{Goldberg1997}. This is impacting organisms underneath and around it, as it suffocates and blocks the light of the coral and soft sediment, which is vital to the ocean ecosystem \citep{Chiappone2005}.\\
Manual efforts have been made to clean up coastlines; Scuba divers such as those who are a part of the Professional Association of Diving Instructors (PADI) Dive Against Debris program, \citep{PADI} have reportedly collected 307,000kg of debris from dives, since 2011. This means that an impressive 2,132 kg of debris is collected a month. Unfortunately, this is problematic as scuba diving is an expensive and hazardous method to rely on, debris collection sometimes requires special training but most notably manual labour is too slow against the rapid influx of debris - the \cite{NOAA2022} have estimated that approximately 666,666 tonnes of plastic alone enters the oceans every month (approximately 8 million metric tonnes a year). This solidifies the reason why automation and artificial intelligence are our best chances of efficiently removing synthetic debris in water bodies, worldwide.
\subsection{The Rise of Plastic}
For almost a century, we have become aware of the scope of repercussions plastic debris is causing and have still not successfully targeted this dilemma.

In the 1930s, a study \cite{Moore2001} found that a small number of northern fur seals had been discovered with “various objects caught around their necks, shoulders and … flippers.” Within the 60s - 70s biologists studying seabird feeding ecology began to notice plastic in their subjects' diets, as well as entanglement in plastic debris in a wide variety of marine organisms. Scientists conducting planktonic and benthic surveys in both the Atlantic and Pacific oceans found unprecedented numbers of plastic particles among their samples \citep{Coleman}.\\
Approximately forty years ago, \cite{Fowler1987} identified that entanglement correlated with the rapid increase of fishing efforts in the North Pacific and Bering sea in the 1960s. This was when plastic materials began to be substantially favoured for creating trawl netting and packing bands.\\
In the 1970s, plastic pollution was found in benthic sediments along the industrialised coast of Great Britain, near Auckland, New Zealand, and in the Mediterranean Sea as enormous floating masses. Plastic pellets were found in coastal regions of the United States, Portugal, Colombia, Lebanon, and remote sites such as the Aleutian and Galapagos Islands. Members of the Marine Resources Monitoring, Assessment, and Prediction Program (MARMAP) found large quantities of raw plastics in the open ocean, particularly in the Sargasso Sea \citep{Coleman}.\\
Later in the 80's, plastic ingestion was observed in a variety of marine animals, including seabirds, fish, sea turtles, and whales. Observations that plastic debris was causing entanglement and death in marine animals, such as gulls and fur seals began to be discovered and the comprehensive study by \cite{Day1980} identified that seabirds eat plastic because they mistake it for natural prey and then discovered that birds could no longer breed effectively.

\subsection{The Plastic Cycle}
Barnacles, kelp \citep{Eriksson2013} and plankton \citep{Moore2001} were discovered with nanoplastics within their systems. These animals are at the very beginning of the food chain. These are the primary food source for filter feeders who are unfortunately also killed by discarded fishing gear in the hundred thousands every year \citep{Laist1997}. Whales are essential to phytoplankton populations; these microscopic creatures single handedly consume approximately “37 billion metric tons of CO, an estimated 40 percent of all CO produced” - the equivalent to 1.70 trillion trees or four Amazon forests \citep{Chami2019}, additionally contributing 50\% of the world’s oxygen within our atmosphere.\\
Due to plastics passing through the food chain and water; \cite{McAdam2017} states that there is no untouched aquatic setting now, and this has even extended to salt marshes \citep{Viehman2011}.\\
With that in mind, it is an epidemic as multiple studies substantiate the consumption of micro plastics  ($<$5mm artificial polymer particles) through fish and seafood \citep{Daniel2020} \citep{Danopoulos2020} \citep{Daniel2021} \citep{Lai2022} \citep{Dong2023}. With degradation, nanoplastics ($<$1$\upmu$m in diameter) are formed \citep{Alimi2018}. Recent research conducted \citep{Leslie2022} found four types of artificial polymers (nanoplastics) in participant’s blood samples including; polyethylene terephthalate (PET), polyethylene (PE), polymers of styrene and poly methyl methacrylate (PMMA). Researchers are still not confident on the adverse health implications this could bring to human life \citep{Smith2018}. \\
In 2015, researchers predicted that by the year 2025, it is estimated that the ocean will have over 250 million tonnes of plastic \citep{Jambeck2015}. Due to the exposure of UV radiation; abrasion and wave action; plastic will break down into micro and nanoplastics  leading to an estimation of approximately 5.25 trillion plastic particles contaminating the sea surface alone \citep{Alimi2018}. Due to their size and irregular properties; these plastics will be difficult to classify and detect.\\
At the time of this survey, a recent study by \citep{Eriksen2023} estimated that within the ocean, at the year of 2019 there were “82–358 trillion plastic particles (mean = 171 trillion plastic particles, primarily microplastics, weighing 1.1–4.9 million tonnes (mean = 2.3 million tonnes)”.\\
As such, the exposure of marine systems to vast (and rising) portions of plastics, including micro and nanoplastics, calls for the benefit of advanced artificial intelligence methods for debris identification.

\section{A Brief History of AI in Marine Debris}
Until 2018, AI for underwater environments and marine debris was hardly researched. This section briefly outlines the use of conventional AI and machine learning models with marine debris before 2018.

\subsection{Timeline}
In 2002, \cite{Foresti2002} used underwater images acquired by an autonomous underwater vehicle (AUV) to develop a hierarchical neural tree classifier (HNTC) for object recognition. The HNTC consisted of macro-pixels and a neural tree that classified each region into different object classes. The results showed that the classification performances remained unchanged in the final images, and the method was slightly slower due to the larger number of pixels to be classified near the borders because of geometrical effects. Overall, the proposed method showed promising results for object recognition in underwater images acquired by an AUV.\\
Later that year, \cite{Boulinguez2002} used acoustic data collected from a parametric sonar to classify buried objects. The study proposed two kinds of methods for selecting key parameters to characterise the 3-D target shape: methods using 2-D projections of 3-D objects and methods using the full 3-D information to compare it with created models. The results showed that the proposed 3-D pattern recognition using decision fusion improved classification rates, however the authors suggested further research to collect data that could help compare methods in a real environment.\\

\cite{Toal} described the development of an intelligent vision system for an autonomous underwater vehicle (AUV) that carries out underwater filming and inspection tasks using a combination of two modules: the video marking system (VMS) and the target tracking system (TTS). The marking system uses different approaches to identify which fragments of the video footage are interesting and discards those that only show the water background. The researchers developed an adaptive vision routine selector that built optimal feature vectors depending on the environment. The results showed that the VMS was successful in detecting interesting events in the video footage, and the adaptive vision routine selector was able to build optimal feature vectors depending on the environment.\\
Two years later, \cite{Walther2004} developed an automated system for detecting and tracking objects in underwater video streams from remotely operated underwater vehicles (ROVs). The researchers used video data from ROVs and implemented a selective attention algorithm to pre-select salient targets for track initiation. The AI used, adopted a saliency-based attention system and object recognition systems. The results showed that the system was able to successfully detect and track objects in the underwater video streams.\\

In 2006, \cite{Mitra2006} used Lidar return data to detect and classify underwater objects. They compared different AI techniques, including parallel neural networks using multilayer perceptron (MLP) and Hierarchical Radial Basis Function (HRBF), Support Vector Machine (SVM) based Information Extraction (IE), Bayesian and quadratic classifiers, and a single level artificial neural network (ANN) architecture. The parallel neural network architecture using MLP was found to have the highest prediction accuracy. The proposed system achieved an overall accuracy of 98.9\%. They concluded that the parallel neural network architecture along with SVM-based IE worked efficiently in detecting and distinguishing the required classes from one another.\\
In the same year, \cite{Cayirci2006} focused on the detection and classification of underwater targets using data mining techniques. The data used was based on the ambient magnetic and acoustic conditions of the vicinity of the targets. The AI algorithms used were one rule (OneR), Naïve Bayes, and J48 decision tree model. The conclusion was that J48 had the highest correct classification rate (92\%) and precision. OneR had a 61\% rate and Naïve Bayes had 84\%.\\

\cite{Balas2006} used data from 157 litter surveys carried out over 49 beaches in Wales, UK to develop an artificial intelligence model for predicting litter categories and grades. They used both a MLP and a fuzzy system as AI techniques. The neural network sub-model could predict litter items in seven categories using general litter data available for a specific Welsh beach, while the fuzzy system sub-model could consider qualitative data described by language, such as questionnaires distributed to beach users. The overall predictions of the number of litter items in seven categories from general litter items at Welsh beaches were considered satisfactory using both AI techniques. The authors concluded that their AI model could save on field effort when fast and reliable estimations of litter categories and grades are required for management studies of beaches.

\subsection{Deep Learning for Other Marine Applications}
Though not related to marine debris, some noteworthy studies have been considered in this section, which we believe may add methodological insight for researchers within the field.\\

In 2004, \cite{Beaufort2004} updated the artificial neural network (ANN) Syste`me de Reconnaissance Automatique de Cocco-lithes (SYRACO) for automatic identification and counting of coccoliths in sediment samples more accurately. Due to the intricate complexity and human error of counting samples, they refined the automated process by using digitised images of sediment samples and trained the ANN using a collection of coccoliths from different species. The new version of SYRACO was tested on a collection set and achieved a correct identification rate of 91\% with six actions. The researchers concluded that the updated version is a robust method for improving the efficiency in the identification and counting of objects in an uncontrolled environment, namely, coccoliths in sediment samples. A limitation expressed by the authors affirms that if a class is underrepresented in a dataset, then the “diffusion of specimens from a more abundant class may cause a significant statistical bias”, articulating the importance of balanced classifications. \\

The study by \cite{Hurtos2013} aimed to develop an automatic detection algorithm for underwater chain links using a forward-looking sonar. The authors communicated the lack of available literature in real time with sonar data. They collected three datasets and had to enhance the data as the accuracy against the original raw files dropped dramatically. Their method showed that the accuracy was “reasonably good” despite the resolution of the data. The conclusion was that the proposed method had high potential to run in real-time and could be used for underwater chain link detection and in this case, cleaning. They elucidate their awareness of their methodology assuming the chain lying approximately on a plane, which they proclaim could be a limitation, though it could be beneficial to explore sonar data underwater further.\\

\cite{Gordan2006} focused on improving the ability to observe object recognition in colour underwater images as such data can often be difficult to analyse due to the conditions. To achieve this, the authors used a SVM classifier. The data was imagery of a hydro-dam wall with the specific intention to localise circular shaped objects of known dimension (the pressure equalisation openings.) The proposed architecture involved extracted pixel features from regions of interest of approximately the object's size. In the classification phase, the underwater image was decomposed into partially overlapping elementary regions of interest and was classified by the SVM based on a shape descriptor (the circularity) of the patterns. This shape descriptor was used for their classification as objects of interest or not, through a simple threshold comparison. The results showed that the proposed architecture minimised the false acceptance and false rejection rates. A limitation of this study could be that the researchers selected only five images of which were augmented into a 255 image database, therefore, it would be interesting to observe the results of this method on a new dataset. \\

These methods have the potential capability to be explored and adapted for the application of detecting debris underwater. For example, methods such as SYRACO present the opportunity to detect and assess complex biological and synthetic structures that humans are prone to making errors with.
Sonar could present an interesting opportunity for application of sonar imaging when detecting basic metallic objects underwater - particularly in deeper depths, where the lighting conditions make image collection and object detection difficult. The authors do specify that as sonar frames cannot detect point features, it would be an unreliable method on optical imagery such as video and photo. However, we question if simple, well structured objects could still be worth investigating. 
Meanwhile, utilising SVM classifiers on marine footage (particularly deep sea) could assist with problems caused by low image quality, low lighting and backscattering from refracted and reflected light on microscopic structures.

\section{Deep learning techniques in marine debris identification}
In order to tackle the ever growing issue of debris accumulating within marine environments; researchers have taken to applying more-advanced artificial intelligence in a variety of methods.  A popular choice has been through the use of remote sensing; with data collected either by drones or through satellites. Remote sensing has had a positive impact on tracking the way floating debris collects or travels with currents.\\
Alternatively, some researchers have taken to applying object detection methods directly on to the surface of the ocean or by applying these methods underwater. Throughout the next few sections, we have organised the research papers within their respective categories in chronological order, so that it is easier for the reader to analyse the data but also to refer back to certain papers.

\subsection{Remote Sensing}
In Chiloe, Chile, anthropogenic marine debris (AMD) has become a significant problem. Researchers \cite{Acuña-Ruz2018} tried to address this by using three different methods of classification when detecting debris on beaches. SVM, RF and Linear Discriminant Analysis (LDA). By using these three models, the researchers created a predictive model for the classification of marine debris within satellite imagery, of which they could compare and select which models were best for specific tasks. Their use of three models also helps to reduce the risk of overfitting to help maintain strong performance levels on new data. SVM performed the strongest for the classification of debris in satellite imagery with an overall accuracy of 80\% when validated against ground truth data. It successfully detected expanded polystyrene and other plastic mixtures on the beach with a spatial resolution between 0.3m to 1.2m and conducted with low errors. The RF and LDA models also performed fairly well with accuracies of 77\% and 70\% respectively. Due to weathering processes the authors found that the spectral signatures of the debris varied widely; for example Styrofoam shows an increase in reflectance by 30\%. \\

The aim of the study by \cite{Jakovljevic2020} was to develop an efficient method for mapping floating plastic. Using two sites, their study is split into two parts; firstly they used UAV images to map where the plastic was but in their second part, they deployed nets to collect the litter. The authors successfully used semantic segmentation pixel classification to detect and map the plastic within water bodies using the ResUNet50 extension. They found that even in shallow waters the model performed well with an average F1 score of 0.89 across the three categories of plastics they searched for. The authors do highlight their agreement with the study by \cite{Ji2015}, who articulate the importance of balanced training sets, as within this research, \cite{Jakovljevic2020} found that mixed pixels would be confused as the majority class. The authors continue their research with their collection of garbage upstream the rivers using nets. They have deduced that annually, the net collects approximately 10,000m3 of materials, including wood (60\%), plastic packaging (35\%) and ‘other’ waste (5\%). This potentially highlights the importance of teaching classification systems to identify wood types and other organic matter that might get washed down rivers but could eventually biodegrade.
\\

\cite{Savastano2021} used Synthetic Aperture Radar (SAR) to detect plastic marine debris in the Balearic Islands. The study trains the AI model on a dataset of pixels, labelled as either plastic or non-plastic utilising three different supervised classifiers; RF; SVM and Gaussian Naive Bayes(GNB). The authors do not make any concluding remarks on the SVM or RF models but did discuss that upon visual inspection; the GNB performed the best on unlabeled images, and that GNB produced a smaller amount of false positives however, when applied against the validation dataset, its performance on labelled areas had the lowest balance accuracy of 0.68 whilst RF and SVM achieved 0.86. The conflicting and inconclusive results of this experiment would suggest that further research on the most appropriate models to be used, would be beneficial, perhaps an application of explainable methods would determine the reasoning behind the level of false positives and provide an opportunity to improve the results. 
\\

In a hybrid method of both tracking and detecting debris; \cite{Kikaki2022} produced a method presenting two weakly supervised semantic segmentation tasks; RF and U-Net. The RF model used spectral signatures of input features (spectral indices, pixel, textural features) extracted from grey-level co-occurrence matrices (GLCM). To take the Sentinel-2 multispectral images as an input; the authors used a U-Net model which output a binary mask that was able to indicate if debris was present or not. To test their benchmark dataset; \cite{Kikaki2022} achieved high accuracy scores and concluded that the MARIDA set is a valuable resource for developing and evaluating ML algorithms, when using Sentinel-2 remote sensing. Their RF model achieved an overall accuracy of 93\%, with a F1 score of 0.68 whilst, the U-Net CNN achieved a high accuracy of 91\%, with a F1 score of 0.60. The performance of the models varied between the classes but their results for the Marine Debris class had a strong outcome. The use of one class to represent all floating debris could cause identification issues due to the vast array of structures and objects this classification represents and it is worth mentioning that with U-Net the marine debris achieved a 0.7 precision-recall average score compared to the best performing version of RF at 0.92.
\\

\cite{Sannigrahi2022} developed a nonlinear kernel Normalised Difference Vegetation Index (kNDVI) for detection of floating plastic in the ocean. They produced two machine learning models; SVM and RF. The SVM method used the kernel; radial basis function (RBF), however, the authors concluded that RF produced more promising results, with a higher accuracy; less errors; and less uncertainty when classifying plastic pixels from satellite data. Within the RF model, the authors used four remote sensing spectral indices; Floating Debris Index (FDI), Normalised Difference Vegetation Index (NDVI), Plastic Index (PI) and (as mentioned above) the kNDVI. The study also conducted hyperparameter tuning such as the number of trees, splits at each node, cost factor and the number of support vectors, which optimised their performance. The authors suggest that slightly weathered plastic objects may cause the machine learning results to fluctuate, particularly as plastic reflects light differently depending on thickness, colour or degradation. Despite this, their RF model still achieved  88\% and 94\% accuracy on their two real world test sites; Beirut and Calabria. They conclude by remarking that remote sensing and machine learning techniques are a plausible combination for detecting marine debris. Based on their research, future studies should take into consideration the way in which plastic and other debris degrades over time, as it is possible this will affect classification as well as tracking results.
\subsection{Classification \& Object Detection}

\cite{Fulton2019} evaluated four popular algorithms Tiny-YOLO, Faster-RCNN (with Inceptionv2), YOLOv2 \& SSD (Single Shot Multi Detector with MobileNetv2) in real time, to compare the results for detection of marine debris - particularly plastics. The authors found that all four models performed well on their image dataset. Their results were also compared on a NVIDIA GTX GPU 1080, NVIDIA Jetson GPU TX2 and Intel™ i3-6100U CPU, to assess real-time performance which varied in frames per second (fps), per model on each machine. Faster-RCNN outperformed the other models in terms of accuracy with a 81\% mAP. The authors comment that they believe Tiny-YOLO and YOLOv2 had a better balance between speed and accuracy; their mAPs were as low as 31.6\% and 47.9\% respectively, their analysis however makes more sense when the plastic detection average precision is taken into account at 70.3\% \& 82.3\% correspondingly, when compared to Faster RCNN’s plastic precision score of 83.3\%, it seems more feasible, as particularly TinyYOLO was significantly faster when using the 1080 GPU, at a rate as fast as 205 frames per second compared to Faster RCNN’s fps of 18.75. Finally SSD scored lowest in terms of detection; 67.4\% mAP with a 69.8\% precision score in the plastic category, it was slowest when using GPUs, however it outperformed the other models in speed when used on the CPU which would operate best on low powered AUVs. Therefore, those who have limited processing power and time may opt in for SSD as an efficient model, though the results would be outperformed by Faster-RCNN. 
\\

\cite{Watanabe2019} explored the use of unmanned aerial vehicles, autonomous underwater vehicles and other autonomous robots to produce a high-spatiotemporal marine-monitoring system. In this study they use the YOLOv3 algorithm for object detection in marine environments. Using this architecture, they found that they were able to detect objects almost in real-time. When tested on underwater sea life, their model achieved a mAP of 69.6\% and underwater marine debris had a mAP of 77.2\%. However, this was built and tested on a visual processing unit (VPU), so they felt they would need to test on various other CPU and GPUs.
\\ 

\cite{Kylili2019}, needed several thousand images to optimise the parameters of their algorithm. As they ran their simulation through training, testing and validation, they required a different set for each. They ran 50 training epochs and achieved incredibly accurate results. The training set was $\approx$100\% accurate with about a 1\% loss, the testing set was $\approx$99\% accurate with $\approx$4\% loss and their validation set achieved 86\% in accuracy.\\

The Visual Geometry Group-16 (VGG16) model made use of transfer learning as it was pre-trained on ImageNet.  A class activation map highlighted the areas within the image that the model was classifying which helped the authors validate their results. \\
Unfortunately, though it is difficult to measure; \cite{Condor} estimates that floating debris only makes up 15\% of the predicted amount of debris in the oceans. Therefore, research depicted on floating debris may not have the desired impact. The authors particularly focused on the credibility of their results (a theme that most authors have omitted within this survey) by reinforcing their results accompanied by the loss percentage as well as altering the structure of their model using three different scenarios.
\\

By applying a Bottleneck (BM) technique to VGG16, \cite{Kylili2020} were able to increase their validation accuracy on detecting floating debris on the Cypriot coastline up to 90\%, as they amplified their dataset and number of classes. They found that even from low resolution images, VGG16 was able to accurately identify marine debris. The authors particularly note the importance of data augmentation to expand the dataset, with that aside, they feel this research could be successfully used for monitoring and managing marine litter. In their conclusion, they suggest applying YOLOv3 in real time to video data. \\
They continued their work \citep{Kylili2021} by collecting shore samples along six different Cypriot beaches and using YOLOv5 and YOLACT++ to classify and localise the debris in the images. From this research they were able to deduce that there would be approximately 66,000 plastic items scattered across Cypriot beaches. They also discovered that the majority of debris sizes ranged from 10cm to 30cm - this potentially offers plausible representation of parameters needed for future researchers. To conclude their findings; they determine the importance of “intelligently [monitoring] litter over large areas of interest in a short period of time” and how that is an integral contribution to marine debris solution.\\
The authors continue to research deep learning methods on debris via both detection and tracking, in a study by \cite{Teng2022}. To estimate the abundance of debris and track objects, they first tracked the data using the centroid tracking technique. They were able to produce a mean average precision of 72\% on a pre-trained YOLOv5 model with the MSCOCO dataset and achieved a classification accuracy of 89\%. \cite{Teng2022} proposed that their method could be used as a more cost effective method to manual surveys for estimating the quantity of debris. They believe their findings could be integrated into other digital applications or other remote methods of surveying, which may enhance its performance. The authors also suggest that this method has the ability to bring the current technology on accurately mapping plastics in the sea, forward to more successful results, which ultimately would help humanity formulate more efficient strategies when addressing any method of conservation efforts within water bodies.
\\

In a study scanning large areas of the sea surface, comparing the results of both Faster R-CNN (FRCNN) and YOLOv5 on their own collected dataset; \cite{deVries2021} found that even though FRCNN used transfer learning with the COCO dataset \citep{Lin2014}; YOLOv5 had a better performance of detecting small objects (YOLOv5: 0.15m, FRCNN: 0.35m). The authors declare that “the detection of the smallest objects is irregular”. To further develop the reliability of their results, they suggest utilising concurrent datasets; additionally the authors expressed the need for gathering further footage. Consequently they disregarded their study as findings but rather as a methodology proposal - though some details on the accuracy of their predictions would have helped with other model comparisons. Their work does however highlight the crucial aspect of the size range of undersea objects - a phenomenon researchers will need to closely observe when collecting and classifying data.
\\

\cite{Deng2021}'s study focuses on the task of marine debris detection and instance segmentation. The authors produce a fortified variation of Mask R-CNN by incorporating dilated convolution in the Feature Pyramid Network, spatial-channel attention mechanism and re-scoring branch. They were able to upgrade the feature extraction and enhance the accuracy of the instance segmentation to a mAP50 of 59.2\% - a 2.5\% increase from the standard Mask R-CNN trained on the TrashCan dataset. Three other algorithms were used to compare the results and they under performed in comparison: 55.9\% (SOLO) and 57.1\% (CondInst) within instance segmentation. Within object detection, their mAP50 reached 65.2\% with improved lateral connection - a 9.5\% increase to the original structure, and again outperforming competing models such as Faster-RCNN (55.4\%), Retina-NET (57.3\%) \& FC0S (60.4\%). Their research particularly exaggerates the importance of reinforcing available models, especially within a specialised, vast field such as marine debris and how the continuous evolution of models is crucial for deep learning development. The downside to this model is its heaviness, limiting its ability to detect in real time but the authors state their intention to develop an optimised, lightweight version.
\\

\cite{Hipolito2021} researched the use of deep transfer learning with the YOLOv3 algorithm, for the detection of underwater marine plastic debris. The researchers utilised a very small sample size of the Trash-ICRA19 \cite{Fulton2020} dataset of underwater plastic waste to train the deep learning system. The study achieved high training and validation accuracy, with an mAP value of 98.15\%. The authors declare their sample size to be small - to produce definitive, conclusive results, their observation is arguably true and the authors do not seem to explain their choice for disregarding the full set to obtain a small sample size. Had they used the full version of Trash-ICRA19, then for a binary classification set of 8,580 images, this study would hold a larger dataset than approximately two thirds of the other papers we have evaluated in this review.\\

\cite{Xue2021a}, \cite{Xue2021b} published two studies in the same year that developed enhanced classification methods to identify and categorise deep sea debris. In one study \citep{Xue2021a}, they proposed a novel network model called Shuffle-Xception. It uses a hybrid combination of depthwise separable convolutions, shuffle operations and shortcut connections to enhance the network's classification performance. The authors compared Shuffle-Xception with ResNetv2-34; ResNetV2-152; MobileNet; LeNet \& Xception. On all seven classes; they found their F1 stop was higher on Shuffle-Xception and their precision scored an average of 95\% suggesting that Shuffle-Xception is an appropriate option for deep-sea debris classification - particularly when considering subject variety and its effects on detection models. In their other publication \citep{Xue2021b}, the authors proposed a one-stage network called ResNet50-YOLOV3; when compared against other detection networks, their results outperformed them in both accuracy and speed with a mean Average Precision (mAP) of 83\%.
\\

\cite{Moorton2022} compared the VGG-16 algorithm with a custom convolutional model on the same dataset which is curated by them. The customised CNN consisted of three convolutional layers with 32 nodes and no dense layer, whereas the VGG-16 model used ImageNet transfer learning weights. Consequently VGG-16 scored 95\% accuracy and the CNN model scored 89\%, however the authors convey the need for a larger database with more diversity.
\\

With a focus on exploring under the surface debris, \cite{Sánchez-Ferrer2022} have produced a dataset suitable for training machine learning models on underwater data. To test the results of their database. They chose to use Mask R-CNN due to its promising ability to detect and segment objects within an image. The model hyperparameters were optimised to improve its performance. The confidence acceptance threshold was modified to 50\% from 70\% which produced a bias of the model towards an overestimation. The authors voiced concerns that the environment conditions of underwater data drastically changed and that it is a challenge to represent the progressive degradation found in underwater data. They tested their model on two underwater videos; one within a controlled fishbowl environment and one within a real-world seabed. They discovered the complexity of the latter negatively impacted their results and consequently, have recommended a greater variety than their data set.\\
Furthermore, the researchers discovered confusion between categories, such as “square can”, “basket” and “metal debris”. These classifications were easily misinterpreted by the framework. Additionally, they found that some categories such as “shoe” were showing large confusion rates due to being underrepresented, highlighting the importance of a balanced dataset. The research also concluded that there were False Positive errors within images that contained more than one single object to detect and that the algorithm favoured the easiest option. They believe this is due to mislabeled objects and where the full object is not visible (partially offscreen or overlapped).\\
The model performed fairly well, with a mAP of 60\% but the authors acknowledge that the size of the CleanSet dataset is limited, does not contain enough variation in debris shapes and colours but also has only been accumulated in a controlled environment.\\
Using Mask R-CNN, \cite{Sánchez-Ferrer2023} used the same dataset with slightly altered classes to progress the recognition performance. The study aimed to explore enhanced techniques that can be used for application on automated vehicles - to achieve this, they fortify their dataset with augmentation and conclude that synthetically produced data is able to boost the overall performance of object detection over the use of real data alone and were able to achieve a mAP of 63.5\% (Instance) and 65.2\% (Material). The research paper does however point out that alternative data generation methods should be explored for more accurate real debris object detection. 
\\

\cite{Zhou2023} aimed to address the challenges of small scale and occlusion of marine debris by developing an object detection network called YOLOTrashCan. YOLOTrashCan consists of two main components; a backbone network which enhances feature extraction and a feature fusion network which combines features from different scales to elevate detection accuracy. The authors found that their model produced a detection accuracy of 65.01\% on the TrashCan-Instance dataset (their Material dataset produced slightly lower results of 55.66\%). They were also able to reduce the network size by 30MB, which makes it a more efficient model for practical applications (such as cleaning marine debris).
\\
In the same year, using the same dataset; \cite{Liu2023} published their findings when they proposed a modification of YOLOv5s on marine debris detection by replacing the backbone with MobileNet and introducing an attention mechanism for filtering key features. Their results found they achieved a 4.5\% mAP increase from the original YOLOv5 model, achieving a mAP of 67\%, whilst simultaneously meeting the requirements of real-time detection. Interestingly, this model outperformed YOLOTrashCan \citep{Zhou2023}, which could refer to the model type but also the increase in classifications that the authors applied to this study.
\\

From the comparison of the available research on detection within marine debris, the results are generally positive with high mean average precision scores. The next section briefly explores similar work on river debris and why it is examined within our survey paper. 

\subsection{Deep Learning for River Debris Identification}

The exploration of addressing river debris is highly important as it is estimated that more than 88-95\% of the input of plastic carried into the sea comes from only ten river systems. \citep{Schmidt2017}, \citep{Jakovljevic2020}. Part of the problem of addressing marine debris is to prevent it from entering in the first place, therefore, a major contribution to resolving this issue is to intercept river debris.\\

\cite{vanLieshout2020} used deep learning to develop a method of monitoring plastic debris in rivers. In their study they used two different CNN’s for different aspects of the research as they wanted to develop the current plastic-waste monitoring methods. To achieve segmentation they used Faster R-CNN. However for the detection, Inception v2 (pretrained on COCO) was trained on the output of the segmentation stage. This helped to detect the regions of the image that contained plastics. The authors logged each of the changes they made to advance the precision score. To begin, they presented the effects of augmentation application on their data. With no augmentation, they strengthened their score of 59.4\% by adding flipping methods which brought their precision rate to 63\%. They then altered their optimiser from momentum to Adam and found their rate reinforced to 65.7\%. Finally, the authors changed their learning rate from fixed to adaptive, which gave them their overall precision score of 68.7\% - a 9.3\% transfiguration from their initial score. \\
The authors found that this research is impacted by the location diversity for the data. To test their method against a ground truth option; they used human counters to monitor their methods and found their technique received similar results. Human labelling reliability can be fickle, which could degenerate the monitoring performance, as humans labelled the 280,000+ objects within the data.\\

Conducted in two different locations; Laos and Thailand. \cite{Maharjan2022} proposed a method of mapping debris on rivers using an autonomous unmanned vehicle by obtaining georeferenced ortho-imagery and combining it with deep learning to detect river plastics. They created two separate databases of floating plastics in rivers, which they trained and tested on four different algorithms from the YOLO family; YOLOv2, YOLOv3, YOLOv4 \& YOLOv5. To robustly test the differences, the authors compared the results of the accuracy, time and computing resources. Interestingly, they found that bright, rigid plastics obtained the highest accuracy and that results of debris covered in sediment suffered; “plastic filled with sand and soil or affected by shadow are ignored”. They also revealed that within shallow waters, the visibility of the water made detection difficult. Atmospheric conditions with data collected from height such as UAV and satellite should be considered to minimise the effects of temperature and wind speed. Overall, the authors suggest optimal weather conditions for high quality data. Due to a discrepancy of light absorption and reflection within differing layers of underwater data - the authors suggest incorporating hyperspectral sensors to help distinguish plastic from other materials. Due to the 83\% mAP and size, they have concluded that the pre-trained YOLOv5s model performs best. The authors released a Phantom 4 drone with 4K resolution camera to sample the distance of approximately 0.82cm to assess the monitoring methods of plastic.

\subsection{Alternative applications}

Applications of 3DCNNs in this field so far have only been used within remote sensing sea ice image classification \citep{Han2020}. Due to the expert knowledge needed within sea ice classification and the manual labelling limitations; this paper is designed to address the problem of small sample sizes. The method called SE-CNN-SVM uses a hybrid network extracting features from hyperspectral images and applies them to a fully connected layer and softmax layer for classification, integrating a squeeze and excitation (SE) block and a support vector machine (SVM) classifier. The results outperformed several classical methods including random forest, decision tree, naive Bayes and k-nearest neighbours; particularly when using small sample conditions. 
\\

Though the research is based on washed up debris rather than within the marine environment, the work by \citep{Fetisov2021}, proves it is still worth exploring, as they were able to determine that a westerly wind was bringing the debris to the Sambian Peninsula in Kulikovo, Kaliningrad, through the use of a decision tree, an unspecified ANN (artificial neural network) and some geographical knowledge (amber deposit wash-outs) This example of research can help us track which tides are carrying debris, if we can track the source then we can implement robust methods of preventing influxes from entering the sea and potentially also have a better idea of the type of debris we are likely to be detecting. The authors conclude that their data is not of a sufficient size for training an ANN but highlight the advantages of these predictions, particularly on anticipating rip tides for coastal regions as they believe it might coincide with marine wash-outs.
\\

The authors \cite{Chang2020} tried to address the growing marine debris concerns on Jeju Island to produce a clean coast detector. They augmented their data to successfully classify their images which they believe is how the Xception model performed so well, as the training set scored 97.73\%. The application is aimed at informing tourists on the cleanliness rate of the coasts to help promote sustainable tourism - as 80\% of all tourism is related to coastal regions and is the lifeblood of many countries (World Resources Institute. 2021); their efforts to reduce marine waste should be considered for tourism and economical conservation.

\subsection{Computational Cost and Performance}

To further understand the speed and performance of some of the models, it is important that researchers disclose the hardware they use. Some authors have shared this information and in this section we have displayed some of the specifications that researchers have provided on Table \ref{table: 1}, to give the reader the option to study the meta data with more of a visual overview. 

\begin{center}
\begin{longtable}{ | m{2cm}| m{6em} | m{1.2cm}| m{6em}| m{13em}| }
\caption{Comparison of models, performance, sample size and hardware use of most recent computer vision models.\label{table: 1}}\\
 \hline
 \textbf{Ref} & \textbf{Model} & \textbf{Data} & \textbf{Results} & \textbf{Hardware}\\
 \hline
 \cite{Acuña-Ruz2018}& SVM \& RF \& LDA & 144 samples & SVM:80\%, RF:77\%, LDA:70\% & PS-300 from APOGEE ® (PS300) for the VNIR range, and TerraSpec 4 Hi Res from ASD Inc. (TS4) for the VNIR-SWIR
range \\
 \hline
 \cite{Kylili2019} & VGG-16 & 12,000 images & Validation accuracy 86\% & Intel Xeon (CPU 2.40 GHz) processor with NVIDIA (Quadro K4200) graphics card \\
 \hline
 \cite{Fulton2019} & TinyYOLO, YOLOv2, SSD, Faster RCNN & 5,720 images & Faster RCNN mAP: 81\% & GPU (NVIDIA GTX 1080); Embedded GPU (NVIDIA Jetson TX2); CPU (Intel i3-6100U) \\
 \hline
 \cite{Watanabe2019} & YOLOv3 & 189 (debris) 8,036 (bio) & mAP Sea life 69.6\% Debris 77.2\% & Intel® CoreTM i7-7800X CPU 350 GHz, 64 bits, 40-GB RAM, GPU Nvidia GTX 1080, CUDA 9.0, cuDNN 7.0.3, and OS ubuntu 16.04 \\
 \hline
 \cite{Chang2020} & XCeption & 3,880 images & 97.73\% accuracy on training set & - \\
 \hline
  \cite{Jakovljevic2020} & ResUNet50 ext of UNet & 3 Datasets: 328; 434; 1,846 images & F1 Score Average 0.89 & DJI Mavic pro equipped with an RGB camera \\
 \hline
  \cite{Kylili2020} & VGG16 & 1,600 raw; 32,000 images & 90\% val accuracy & Intel® Xeon® machine equipped with an Intel CPU Core E5-2630 v3 (2.40 GHz) with 48.0 GB of memory (RAM) and an NVIDIA Quadro K4200 graphics card, clock-rated at 784 MHz with 28.6 GB of memory \\
 \hline
 \cite{Panwar2020} & RetinaNet (Resnet50 backbone \& FPN) & 369 images & mAP 81\% & - \\
 \hline
 \cite{vanLieshout2020} & Faster R-CNN and Inception v2 & Raw 1,272 jpg & Precision: 68.7 & -
 \\
 \hline
\cite{Han2020} & SE-CNN-SVM & 3,190 samples & 97.42\% & Intel Core i5-4590 with 3.30 GHz and Nvidia GeForce GT 705 \\
 \hline
\cite{Savastano2021} & RF; SVM; GNB & 8,395 pixels 1,794 plastic pixels & Balanced acc: 0.86;0.86;0.68 & - \\
 \hline
   \cite{Kylili2021} & YOLOV5 \& YOLACT++ & 1,650 images & AP 92.4\%; 69.6\% & NVIDIA® Tesla® K80 graphical processing unit (GPU) \\
\hline
 \cite{Xue2021b} & ResNet50-YOLOv3 & 10,000 images & mAP 83.4\% & GeForce GTX 1080Ti GPU with- a capacity of 11 GB \\
\hline
  \cite{Deng2021} & Mask RCNN & 7,212 images & mAP 59.2\%; 65.2\% & Ubuntu18.04. Intel(R) Xeon(R)Silver 4110 CPU @2.10GHz. GeForce RTX 2080Ti. CUDA 10.2. \\
 \hline
 \cite{Xue2021a} & Shuffle-Xception & 13,914 images & 95\% precision average & GeForce GTX 1080Ti GPU and installed with Intel(R) Xeon(R) W-2133 CPU at 3.60 GHz, 31.7 GB RAM \\
 \hline
 \cite{Hipolito2021} & YOLOv3 & 300 images & mAP 98.15\% & - \\
 \hline
 \cite{Fetisov2021} & ANN & 25 cases & - & - \\
 \hline
 \cite{deVries2021} & Faster R-CNN-Ultralytics YOLOv5 & 18,589 images & - & GoPro Hero 6 Black \\ 
 \hline
 \cite{Teng2022} & YOLOv5 & 2,050 images & mAP 89.4 \% & NVIDIA® Tesla® K4 GPU from Google Colab \\
 \hline
 \cite{Maharjan2022} & YOLOv2; YOLOv3; YOLOv4; YOLOv5s & $>$500 tiles & mAP 77\%; 81\%; 83\%; 83\% & (1) Anaconda with Jupyter running on a personal computer with an Intel®CoreTM i7-10750H CPU @2.60 GHz, 16GB RAM, and NVIDIA GeForce RTX 2060 GPU with 6 GB GPU RAM, and (2) Google Co-laboratory Pro. The personal computer was used for YOLOv3 and YOLOv5, and GoogleColaboratory Pro was used for YOLOv2 and YOLOv4 \\
 \hline
 \cite{Sannigrahi2022} & SVM \& RF & 27 Sentinel-2A/B scenes & 88\% \& 94\% accuracy & - \\
 \hline
 \cite{Moorton2022} & VGG-16 \& custom & 1,744 images & accuracy of 89\% CNN and 95\% VGG-16 & Dell Inspiron i7-7700HQ CPU 2.8GHz, 16GB RAM, 64-Bit with a Nvidia GeForce GTX graphics card, running on Windows 10 \\
 \hline
 \cite{Kikaki2022} & Random Forest model (RF) \& U-Net architecture & 1,381 patches; 3,399 marine debris pixels & 92\%; 70\% accuracy & - \\
 \hline
 \cite{Sánchez-Ferrer2022} & Mask R-CNN & 1,223 images & mAP 60\% & Intel(R) Core(TM) i7-8700 CPU @ 3.20 GHz with 16 GB RAM, a NVIDIA GeForce RTX 2070 with 6 GB GDDR6 Graphics Processing Unit (GPU) with the cuDNN library \\
 \hline
 \cite{Sánchez-Ferrer2023} & Mask R-CNN & 1,223 images & mAP Instance 63.5\%; Material 65.2\% & - \\
  \hline
 \cite{Liu2023} & YOLOv5 modified & 7,212 images & mAP 67\% & Intel(R) Xeon(R) Silver 4210R
CPU@2.20GHz with an NVIDIA GeForce RTX 2090Ti GPU \\
  \hline
 \cite{Zhou2023} & YOLOTrash-Can & 7,212 images & mAP 58.66\%; 65.01\% & AMD Ryzen 7 3700X, Nvidia TITAN RTX 24 GB, and RAM 48G \\
   \hline
\end{longtable}
\end{center}

\section{Data Acquisition}

Conclusions across various marine debris studies, display a common recommendation as researchers and scientists declare that there is not a public database available to solidify their results. Therefore, this section of this survey explores the various methods of collecting data, augmenting, labelling and classifying, that scientists in the previous five years have used.

The JEDI (JAMSTEC E-Library of Deep sea Images) dataset will get mentioned a few times within this section. It is a part of the JAMSTEC (Japan Agency for Marine-Earth Science and Technology) data collection and is popular in many studies, as so far it has been the most comprehensive open access collection of sea debris and biodiversity footage underwater \citep{JAMSTEC}. These deep sea photos and videos are taken from a submarine off the Japanese coast with varying conditions. However, the downside to this database is that it is available for multiple uses rather than specifically for machine learning. Therefore, there is a lot of data cleaning - some images or footage contain parts of the ROV system in place, the quality is not always to a computer vision quality standard and the data would need to be labelled appropriately. Hence also why this dataset has not been included in \textit{4.1 Benchmark Databases} and alternatively, is mentioned here.\\

When detecting debris on the beaches of \cite{Acuña-Ruz2018} used an interesting method of collecting data through fieldwork, lab analysis and satellite data. To produce a ground truth dataset, as well as a dataset used to create a predictive model for the classification; the fieldwork collected 144 samples of various macroplastics and organic material. All of the samples were characterised, labelled and where possible, geo-referenced. These were moved to be analysed in the lab later. Within the laboratory the team were able to observe the spectral signatures of the collected samples using a hyperspectral sensor. From each class; the researchers selected 90 pixels from the scene to develop a supervised classification predictive model which was split at a 70:30 ratio. The other dataset consisted of satellite imagery with a spatial resolution of 0.31m to classify the debris on the beaches they were observing. Due to atmospheric effects on satellite imagery (such as weather), the authors were able to use DigitalGlobe to correct the effects.\\

\cite{deVries2021} collected 4,000 images with a GoPro attached to a vessel. They applied random transformations including “image shear along the X and Y
axis; zoom in/out; rotation by an arbitrary angle; slicing of images to produce different
scales; translation; random exposure and contrast [and] multiplication of the digital number
(DN) values”. This resulted in a dataset yielding 15,227 images. For such a large database, they sought out an online labelling platform called Zooniverse \citep{Zooniverse2023} which the authors chose to categorise into three classes: hard plastics, nets, and ropes - this data was was cleaned using a VGG annotator and they were careful to only label clearly identifiable debris. Afterwards, the data was combined into one class of ‘debris’ as the authors claim it improved the detection precision. The decision to limit false positive detections by including an additional 3,362 photos of the sea without any debris which exhibited a variety of sea states; brought their training set to 18,589 images and their validation set of 739, what’s more, the extra data functioned as a means of quality control.\\

J-EDI was used for research by \cite{Fulton2019} who collected 5,720 images drawn from video data and annotated them with the LabelImg software. Their dataset Trash-ICRA19 \citep{Fulton2020} was divided into three classifications including Plastics, ROV (containing all synthetic objects intentionally placed) \& Bio and their test set consisted of 820 images. Due to the overwhelming assortment of marine debris, they chose to focus on plastics as it is the most harmful material in the ocean currently. Regardless of the difficulty of obtaining this data, their set - of which they describe to closely mimic the real world - has a high standard of environmental variety factored in, due to the range of light conditions, overgrowth, states of decay and occlusion.

Using a GoPro Hero 6 and scuba diving off the Japanese coast, \cite{Watanabe2019} recorded a 6m visibility at 12m depth noting that it was cloudy, to collect imagery of underwater floating debris. They also used three cameras to photograph debris along the beach as well as downloading imagery from Google Open Images. Using a different detection model for both marine biodiversity and debris: for three categories of fish, turtles and jellyfish; they used 6,908 images for the training and 1,127 for testing at roughly a 86:14 ratio split. The neural network used for debris split the data into plastic bottles, plastic bags, driftwood and other debris; due to unavailability of obtaining much underwater data; this model was trained with a much smaller collection of 189 images and just 37 images for testing - a better ratio (80:20) but room for bias and consequently the set too small for debris detection.

Much like the previous work mentioned, the set produced by \cite{Chang2020} collected a small dataset that would not provide enough information for computer vision training. However, their method of collecting data was to select and download 194 Google images which they split into the binary classification of ‘clean’ and ‘polluted’ beaches. Due to the small size of the database, after removing irrelevant files that overlapped; the authors labelled 97 training images and used 20 images within the validation set. To test the model they used the remaining 10 images to conclude their overall database at a quantity of 127 images.

Spanning two sites in Lake Balkana; Crna Rijeka and Vrbas Rivers, the authors \cite{Jakovljevic2020} collected high resolution orthophoto image data. Using a DJI Mavic pro equipped with an RGB camera, six AUV surveys at a height of 12 - 90 metres were conducted. To preprocess the images into a higher quality, the authors used the Structure from Motion (SfM) algorithm. The floating debris most analysed were three variants of plastic; OPS, Nylon and PET. The authors ability to track floating plastic could be a beneficial addition to monitoring debris trajectories in the ocean.\\

When training their dataset for river plastic monitoring; \cite{vanLieshout2020} captured a dataset of objects floating on the river by camera. The authors used the data to train a CNN which organised the images into a binary classification of either plastic or non-plastic. To collect their data; they recorded a video of floating debris on the river for 26 days; of which 1,272 JPEG images were extracted to create the river image dataset. These were collected over five different locations. To create the floating plastic dataset; they uploaded their set to be labelled by citizen scientists on Zooniverse \citep{Zooniverse2023}, which were later inspected and corrected where necessary. Out of the labelled 280,832 objects, 14,968 boxes were floating plastic. The data was split into their binary classification and then was further divided into five subsets of the locations they shot their footage at.\\

Marine \& Carbon Lab, are the organisation in the University of Nicosia who have co-produced multiple studies in the field of detecting marine debris and were able to pretrain their VGG16 model on ImageNet and then continue training the model with their own previously curated dataset of floating marine debris as well as some donated images from a nonprofit organisation called Algalita (https://algalita.org/). In their 2019 study, each of the three categories of floating debris; plastic bottles; plastic buckets; plastic straws contained 250 images in the research; however, data augmentation in the form of zooming, rotating, shifting and flipping increased this number to 4,000 images each. The set was then split into 3,200 train images and 800 for testing. The next year, \cite{Kylili2020} separated their data into eight classes; six for synthetic debris which included ‘plastic bottles’, ‘plastic buckets’, ‘plastic bags’, ‘fishing nets’, ‘plastic straws’, and ‘food wrappings’. Additionally, they chose to integrate a class of ‘Flying Fish’ and also an ‘Other’ category. Each class contained 200 images, using augmentation, they manipulated each image 25 times to create a class of 4,000 images and an overall dataset of 32,000 images; the classes were then split to an 80:20 ratio for testing and training.\\
Following this research, the team and another researcher further developed their study. \cite{Teng2022} collected a database consisting of 2,050 videos split between nine categories of plastic bottles, plastic buckets, plastic bags, fishing nets, plastic straws, food wrappings, a fish species, aluminium cans and cigarette butts. The data was collected from similar sources such as ImageNet and Algalita, but their YOLOv5 model was pre trained on MSCOCO. To annotate their database the authors used the YOLO\_mark and Make-Sense tools \citep{make-sense}, which were used as a method to manually draw the ground truth bounding boxes. The authors declare that their data is available upon request.\\

\cite{Hipolito2021} obtained their data from the Trash-ICRA19 dataset \cite{Fulton2020}. The authors collected a small sample size of 300 images split into the 80:20 ratio and relabelled the set to Non-Bio to keep it as a singular classification system. \cite{Hipolito2021} claim the original dataset held two classifications; bio and non-bio split equally with 4,290 images per category, totalling the collection at 8,580 images, however at the time of writing this paper, we found that the same dataset currently consists of 5,700 images categorised into three main classifications and the original paper referenced \cite{Fulton2020} states 5,720 images. We are not confident if this is a citation error or if the original dataset has since been adapted in some way such as employing further data cleaning and/or data augmentation techniques.\\

Between the two studies; \cite{Xue2021a} \& \cite{Xue2021b}, collected a relatively similar database from the underwater open source collection that is provided by the Japan Agency for Marine-Earth Science and Technology (JAMSTEC). The authors extracted the appropriate frames they needed from the archive of footage with only a single object within each frame, using ‘LabelImg’ \citep{Tzutalin2015} to label their dataset of which they have named 3-D dataset (not to be confused with three dimensional); this particular collection of data consisted of 10,000 images \citep{Xue2021b} - their other study consisted of 13,914 \citep{Xue2021a}. Both databases were organised into seven categories; cloth, fishing net \& rope; glass; rubber; plastic; natural debris and metal - though the training set is comprehensive; the classes are unbalanced in \cite{Xue2021a} ranging from the smallest sample size : 1,825 (Natural Debris), to the largest: 2,234 (Fishing Net \& Rope). The DDI dataset was divided into three subsets; 70\% training, 15\% validation and 15\% testing.\\

Though the amount of data collected is not specified in the paper; \cite{Maharjan2022} used the YOLOLabel tool to record the bounding box for each piece of plastic in each image they had collected of river debris; this resulted in annotating 500 tiles for each river. The labelled objects were then classified into a binary group of either plastics or non-plastics.\\
By contacting a variety of organisations and dive schools around the globe; \cite{Moorton2022} were able to collect and label a small database of 1,644 underwater images using augmentation and 100 testing images. The dataset exhibited diversity in lighting and environmental factors; it also had a large scope of biodiversity as well as synthetic debris, however the set was split into a binary system consisting of Animal (which depicted all biodiversity) and Litter and also was not split at the standard 80:20 train/test ratio - the authors found that due to their dataset, the models were making similar mistakes to turtles when it came to differentiating between jellyfish and plastic bags or wrappers. The dataset is available upon request from authors.\\

\cite{Sannigrahi2022} collected their data from two sources. For in-situ plastic locations; they collected data from \cite{Topouzelis2019}, \cite{Topouzelis2020} in Greece and \cite{Themistocleous2020} in Cyprus spanning from 2018, 2019 and 2021. Using a three class classification scheme; the data was split into Plastic Object; Natural Debris (wood) and seawater. Later, the authors also conducted a fourth class to detect mixed objects - both plastic and wood. The remote sensing was able to manually detect floating debris using the FDI. To collect their Multispectral Imagery, they were able to use data from Copernis Open Access Hub \citep{ESA}; this database was used for the remote sensing analysis. They collected 27 Sentinel-2A/B scenes which were processed with the European Space Agency’s software; Sentinel Application Platform.\\

\cite{Sánchez-Ferrer2023} followed their previous work based on the CleanSea dataset, where they have collected 1,223 images for seventeen classes including: Plastic bag, Towel, Can, Glove, Packaging, Basket, Bottle, Pipe, Fishing net, Metal waste, Plastic waste, Square can, Rope, Tire, Bumper, Shoe, Wood. The set was organised in an 80:20 split with 10\% validation. They opted in for augmented data to amplify the results. The augmentation demonstrated a combination of transformation methods randomly applied to each image, methods incorporated were “horizontal flips, mirroring effects, Gaussian blurring, contrast alterations, brightness modifications, scale changes, translations, image rotations, and image shearing”, the random application was an intuitive way of preventing overlearning. A large increase  via data augmentation provided drastically better results.

\cite{Liu2023} used a dataset of 7,212 images collected from the TrashCan-Instance and split their data into 50 class names (of which were not elaborated on within the paper); though the individual classification are not specified; the authors describe the categories as underwater observations of trash, ROV’s as well as a variety of flora and fauna. To have such a large number of classifications is an exciting step forward in this heavily underrepresented area of research; however, assuming classes are balanced, it would mean that each class only contained approximately 144 images. This therefore, would still not be a quantifiable amount for computer vision training. Though this is not clearly described within the paper and can be assessed more closely on the TrashCan-Instance dataset produced by \cite{Hong2020}.
\\

To conclude this section; data augmentation has been instrumental in producing results and based on the methodology conclusions, we know that the use of augmentation produces larger datasets that are less likely to be over fit or obtain bias. Unfortunately, the amount of data and classifications with enough data within them, still limits the available research, leading to inconclusive and questionable results. \\
Though many researchers provide examples of the imagery they have used, unless otherwise stated within their respective sections; to the best of our knowledge, most research papers did not provide access to their data or we were unable to locate them. Hence there is still an open gap of public data availability in this field. 

\subsection{Benchmark Databases}

This section briefly outlines some of the benchmark databases of underwater debris imagery that researches have collected from.

\cite{Kikaki2022} produced a multispectral Sentinel-2 satellite database of geo-referenced pixels called Marine Debris Archive (MARIDA). MARIDA was designed to produce a suitable, labelled data collection as a benchmark for anyone wishing to build or evaluate a machine learning algorithm for marine debris detection. The researchers take into consideration the importance of geographically diverse data and additionally have built their set using eleven different categories consisting of sea surface debris, biodiversity, ships, clouds and water states. The set was split into a train, test and validation set as a 50:25:25 ratio. Rotations and horizontal flips were used to further expand the dataset with augmentation, bringing their overall dataset quantity to 1,381 patches, that included 837,357 annotated pixels, of which 3,399 were marine debris. Uniquely, their dataset is able to distinguish between various marine features that coexist; such as \textit{sargassum microalgae}, ships, foam, and other natural organic materials. The authors focused on the annotation regulations with precision by drafting three experts to label the set, and an inter-annotator agreement to regulate the annotations.
\\

TACO (Trash Annotations in Context for Litter Detection) produced by \cite{Proença2020} is an open source database that collects annotated images of urban trash. \cite{Panwar2020} adapted some of this database for marine debris detection by producing the dataset AquaVision which consisted of 369 underwater annotated images from TACO. The authors of AquaVision categorised the images into four: glass; metal; paper; plastic. Though the sample size is small, the authors still obtained a mAP of 81\% - even when tested on more random images. The authors claim the small dataset provides “results in a more efficient manner”, however due to the size and limited categories, it would be interesting to test how the model and data would perform on a larger scale with more complex tasks such as avoiding misidentification between plastic and jellyfish. 
\\

The dataset TrashCan used by \cite{Zhou2023} was created by \cite{Hong2020} pulled from JAMSTEC videos and the authors collected 7,212 underwater RGB stills. The TrashCan dataset is separated into two parts; \textit{Instance} and \textit{TrashCan-Material}. \textit{Material} contains sixteen categories such as animal\_crab, plant, trash\_fabric, a full list is given in Table \ref{table:2}. The \textit{Instance} dataset contains the same categories but has removed the ‘trash\_’ sub-categories and altered them to include a finer list of categories totalling twenty two classifications as listed under the header \textit{‘TrashCan-Instance’} in Table \ref{table:2}.  As we can see from comparing the lists, the \textit{Instance} dataset details their trash categories in more depth and ultimately that could be the result of why its performance is stronger than the \textit{Material} set. The \textit{Instance} set was also used in the study \citep{Deng2021}, to improve their model of Mask R-CNN in object detection and instance segmentation. They also applied data enhancement such as image rotation and cropping to ensure effective feature extraction but the overall set number was not disclosed.

\begin{table}[h!]
\caption{TrashCan dataset classifications}
\begin{center}
\rowcolors{13}{gray!50!white!50}{grey!70!yellow!40}
\begin{tabular}{ |c|c|c|c| }
 \hline
 \multicolumn{2}{|c|}{TrashCan-Material} & \multicolumn{2}{|c|}{TrashCan-Instance} \\
\hline
 animal\_crab & trash\_metal & animal\_crab & trash\_can
 \\
 \hline
   animal\_eel & trash\_paper & animal\_eel & trash\_clothing  \\
 \hline
   animal\_etc & trash\_plastic & animal\_etc & trash\_container  \\
 \hline
   animal\_fish & trash\_rubber & animal\_fish & trash\_cup \\
 \hline
   animal\_shells & trash\_wood & animal\_shells & trash\_net \\
 \hline
  animal\_starfish &   & animal\_starfish & trash\_pipe  \\
 \hline
   plant &  & plant & trash\_rope  \\
 \hline
   rov &   & rov & trash\_snack\_wrapper \\
 \hline
  trash\_etc &   & trash\_bag & trash\_tarp \\
 \hline
   trash\_fabric &   & trash\_bottle & trash\_unknown\_instance  \\
 \hline
   trash\_fishing\_gear &   & trash\_branch & trash\_wreckage \\
 \hline
\end{tabular}
\label{table:2}
\end{center}
\end{table}

The authors of this research used 6,065 images for training the \textit{Instance} category with 1,147 in the validation set. On their separate set for \textit{Material}, their train set consisted of 6,008 images and 1,204 validation. The authors believe this provided a comprehensive representation of marine debris and other objects present. They noted that many of the marine debris categories had similarities in the object properties but did not disclose what the categories were.\\

Another benchmark marine debris dataset was produced by \cite{Sánchez-Ferrer2022}. The CleanSea dataset is composed of submerged debris and mostly collected from Marine-Earth Science and Technology (JAMSTEC) data which are videoed and photographed from submarines in Japan. From JAMSTEC, \cite{Sánchez-Ferrer2022} selected 1,223 images, which were annotated by bounding boxes and contours of objects and then were organised into 19 classifications of debris and tested on a Mask RCNN model.
\\

Although there are attempts of curation and use of image databases with either floating or underwater marine debris scenarios, the repositories publicly available are still too undiversified and small to produce the results that researchers are looking for. Hence curation and making available a benchmark dataset that is open to all researchers would address one of the most common obstacles for deep learning in this field.

\section{Survey Overview and Conclusion}

To combine all the metadata of the papers reviewed in this survey paper, \textit{Supplementary Table 1} can be used to cross reference the model performance, amount of data, classification groups and even the hardware to determine more accurately which results have produced the most reliable outcomes.\\
Across the available studies we have observed that there is a major lack in availability of data for marine debris whether underwater or on the coastal beaches. Therefore, we decided to test to see if a small dataset could still produce reliable results on the most successful method of detection that we observed across the research.\\

\begin{figure}[h!]
    \centering
    \includegraphics[width=45mm]{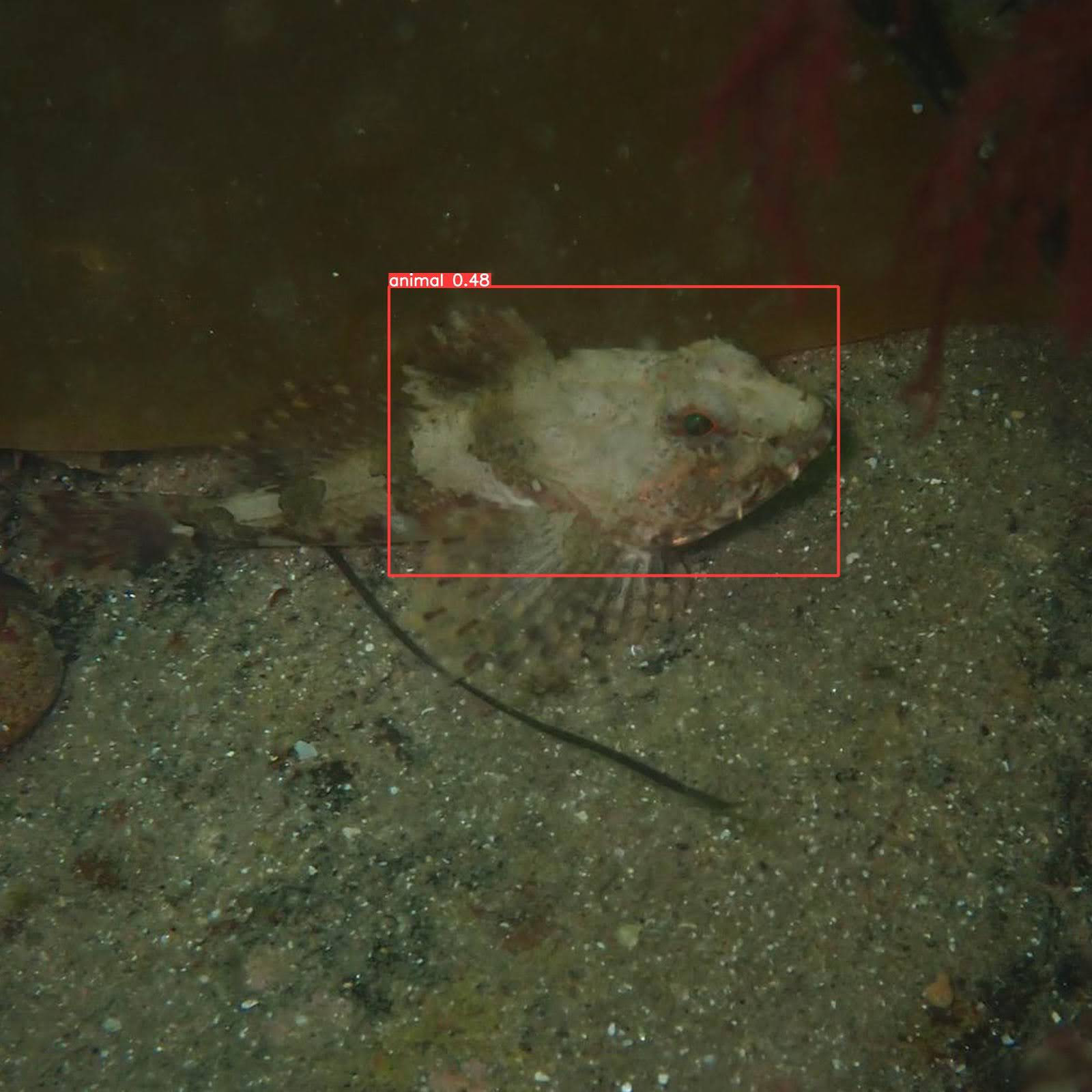}
    \includegraphics[width=45mm] {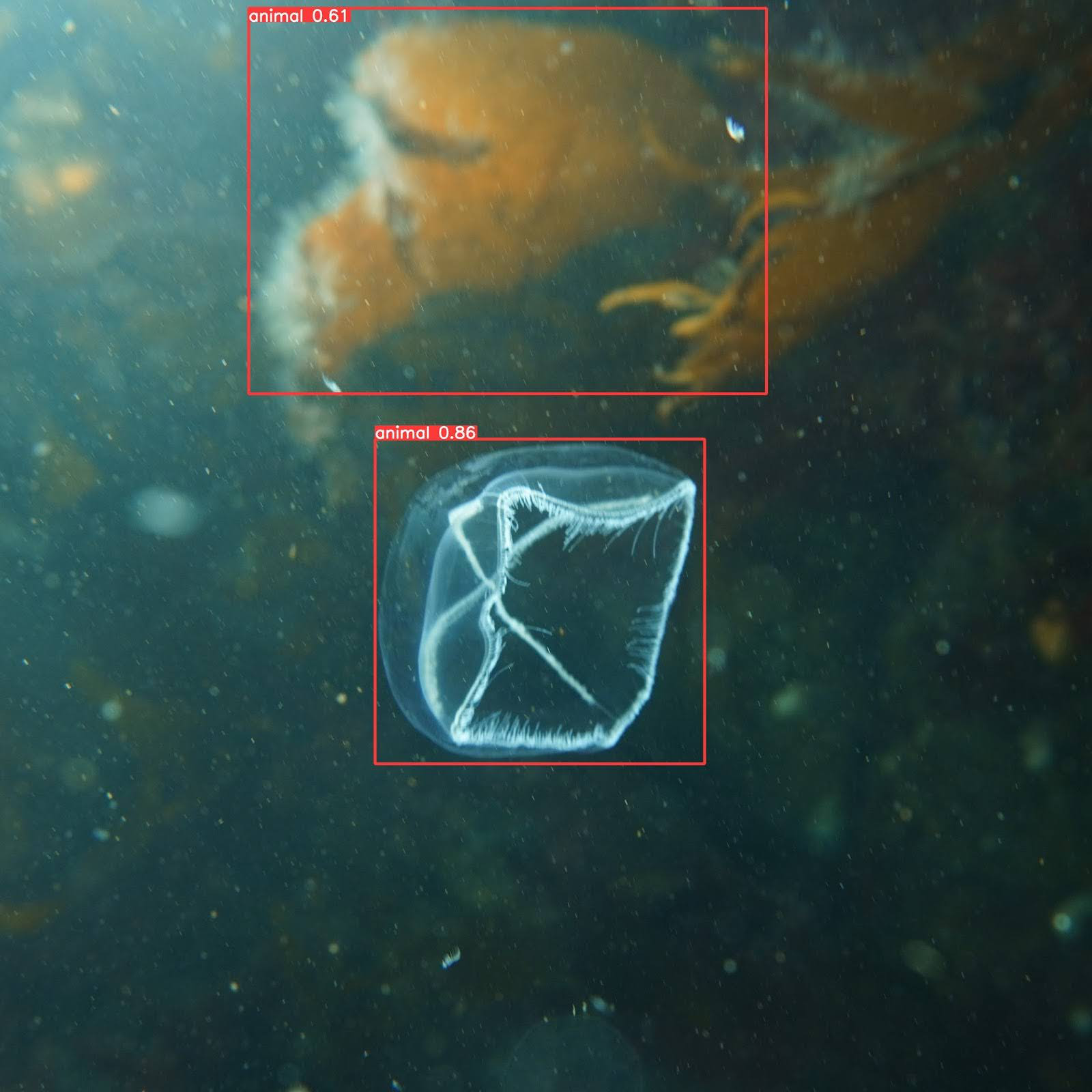}

    \includegraphics[width=45mm]{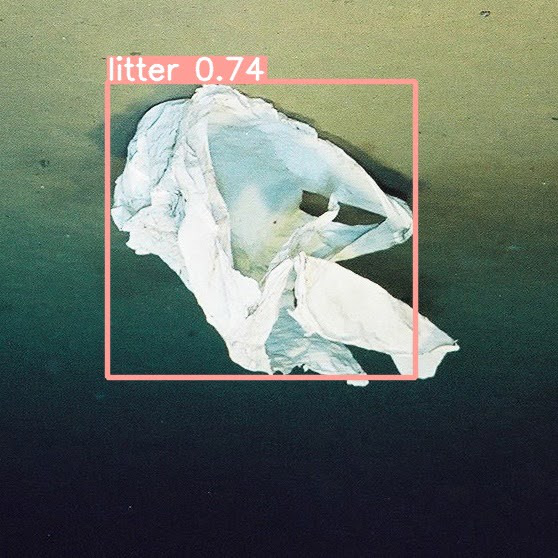}
    \includegraphics[width=45mm]{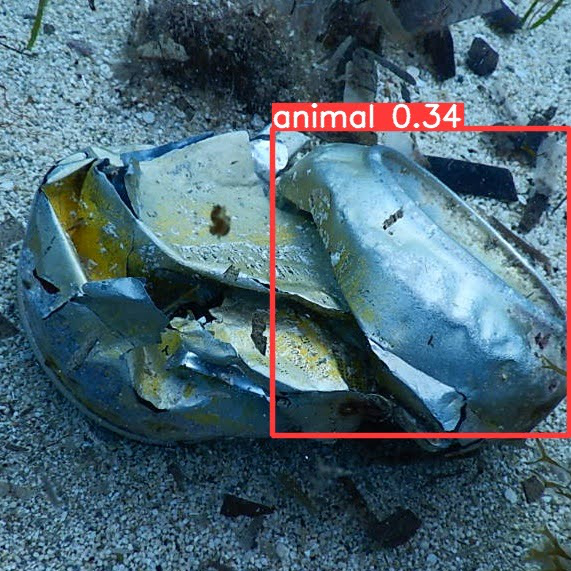}
    \caption{Outcome of highest accuracy test. Figures A and B (Top Row): Animal. Figures C and D (Bottom Row): Litter.}
    \label{fig:Figure 1}
\end{figure}

There is a strong correlation between YOLOv5 and a high mean accuracy precision. Therefore, we decided to test YOLOv5, as it is one of the most prominent methods highlighted in the relevant literature as presented in the previous chapters, (Git pulled from \cite{Singla}) on our small humble dataset to directly compare the results. Our dataset was only a small binary collection of 140 testing images (70 for each category and an 80:20 validation split). We chose a binary classification of litter and animal, as two classifications was the most popular choice across the survey and we wanted to represent as closely as we could. We trained the data using batch sizes of 16, 32 and 64 and epochs of 50, 100 and 150. The optimal parameter settings was found to be a batch size of 32 and epoch number of 50 yet we found that with YOLOv5 we were unable to accurately detect the two classifications; there were many false positives and negatives however most images including object such as plastic bags were unable to be detected at all.\\
As this dataset was limited, we decided to increase the amount of data and attempt to retrain YOLOv5, to check if it was any more useful. Unfortunately, even with a significantly larger dataset of 1,093 training images, 269 validation and 30 brand new testing images (with 38 objects, two example images from each class were shown in Figure \ref{fig:Figure 1}), we were not able to obtain favourable results, as shown in Table \ref{table:3}, leading us to believe that a small dataset is not plausible for use in this area of research. We therefore highlight that a diverse variety of data and classifications is crucial to the successful training of object detection on marine debris.

\begin{table}[h!]
\begin{center}
\caption{Results of testing our custom training set on YOLOv5}
\label{table:3}
\begin{tabular}{ |c|c|c| }
 \hline
 Batch Size & Epoch No. & Accuracy of Classification of 38 Objects \\
 \hline
  16 & 50 & 52.63\% \\
  \hline
  16 & 100 & 44.73\%\\
 \hline
  16 & 150 & 47.36\%\\
  \hline
  \textbf{32} & \textbf{50} & \textbf{63.15\%} \\
  \hline
  32 & 100 & 42.1\%\\
 \hline
  32 & 150 & 52.63\%\\
   \hline
  64 & 50 & 52.63\% \\
  \hline
  64 & 100 & 39.47\%\\
 \hline
  64 & 150 & 60.52\%\\
 \hline
\end{tabular}
\end{center}
\end{table}

\section{Discussion and Future Direction} 

In order to provide future work suggestions, we have collected other authors’ recommendations within this survey, as well as concluding with our own ideas accumulated from this research. This section has been organised into various categories for easier reference.

\subsection{Data Collection Refinement}

The collection of data within the marine debris topic has proven to considerably affect the results that researchers are claiming. With many authors stating that database collections are just not large enough \citep{Maharjan2022}, \citep{vanLieshout2020}, \citep{Zhou2023} particularly of underwater debris, it would make sense for a large database to be produced. Potentially even the use of multiple databases could be explored, as it is limited in this area of research.
\\

\cite{deVries2021}’s research attempted to produce a comparative study between YOLOv5 and Faster RCNN but the authors decided their dataset was not large enough; it would be beneficial to obtain a sizeable increase in images and train the methodology the authors have recommended. Individually the models have produced high mean average precision scores when detecting debris, accordingly, a comparison would be a valuable insight.
\\Furthermore, comparing these object detection results on the same dataset as classification methods which have also performed well such as VGG-16 and ShuffleXception would be a meaningful study to determine which method of computer vision is superior for sea debris detection.\\

In addition to an increase in image availability, researchers such as \cite{vanLieshout2020}, \cite{Hipolito2021},  \cite{Maharjan2022} and \cite{Zhou2023} have pointed out that the variation of images is also not up to a high enough standard. Their suggestions range from diversifying weather conditions, locations and visibility within the water. A few authors request more optimal conditions based on a higher accuracy but in order to provide an AI framework that works in multiple conditions, perhaps the databases should embody more challenging pieces of data to enrich the robustness of the model. 
\\

With the dilemma of debris quantity and variation ever increasing, a high number of classes should also be available to the dataset, to broaden the complexity of the model.\\
\cite{Jakovljevic2020} found that approximately 60\% of the debris they catch in rivers is a wood product or similar; it could be valuable to ensure that classification systems also include wood materials, to increase accuracy rates on current models. We believe it would be highly beneficial to train river classification models on wood types, to prevent mistakes in detection but also if the algorithm were to be applied to a form of automation, collecting wood would not need to be a priority as it is mostly a natural and biodegradable material but it might also help to eliminate precious resources - with that in mind; potentially harmful types of manufactured woods or those with coatings could be an area of classification to explore. \cite{vanLieshout2020} suggest that future research should monitor the reliability of human labelling; increasing the precision of how we categorise data.

Within their research exploring the use of SAR to detect debris, \cite{Savastano2021} highlighted the importance of using a reliable ground truth dataset within AI training. They recommended researchers attempt to strengthen the ground data by focusing on the use of in-situ validated ground truth data in a diverse range of locations.\\

The authors \cite{deVries2021} found that their method cannot currently fill the gap of debris from 50cm down to 5cm, due to the size being too small to be detected. As most plastics continue to degrade over time, they recommend future work in this area.
\cite{Liu2023} also highlighted the importance of small object detection; they suggest improving network structure or even exploring alternative techniques.

Within macroplastics there is rich literature on the applications of AI but a microplastics and nanoplastics review study could be explored. It is an extremely complex challenge and would require an intricate detection system, with data collected from under a microscope, perhaps the classes could be separated into colours or material properties.\\

In addition, the condition of any size debris should be considered, after proving that brighter and more rigid pieces of plastic perform well; how would we solve degradation of colour and texture within debris underwater and exposure to natural elements? Furthermore, it would be interesting to observe the results of detection within debris partially or fully covered in sediment or wildlife. Based on the findings from \cite{Sánchez-Ferrer2022}, it would be valuable to explore research that is able to either detect more than one object within a frame or be able to explore partially hidden objects.\\

\cite{Sánchez-Ferrer2023} suggested a plethora of future works such as upgraded synthetic data generation to better replicate real ocean conditions. They suggest the following considerations: “object integration, scaling policies, and depth simulation”.
Secondly, they recommend domain adaptation to reduce the need for annotated data. They also suggest exploring the use of GANs (Generative Adversarial Network) for generating synthetic data, of which \cite{Fulton2019} also expressed interest in. Finally, based on their conclusions, they recommend a deeper exploration of the integration of both real and synthetic data, potentially investigating the optimal percentage of synthetic data that advances the overall performance of the model.

\subsection{Monitoring Debris}

Various remote sensing applications could be further modified to conduct long term monitoring systems, which could enable the prevention of debris entering large water bodies.
Additionally, the assessment of temporal dynamics of floating debris and its interaction with environmental factors could be a valuable analysis of debris behaviour in water flow and weather conditions. A study by University of Oxford; \cite{Vogt-Vincent2023} was able to source the terrestrial and marine origins of debris into the Seychelles islands using the Lagrangian trajectory - some of the floating plastic was believed to travel anywhere up to six months to two years. It could therefore be interesting to employ machine learning methods to track where these items may have come from; helping to reduce initial improper waste disposal.\\
Contributions of citizen scientists, as \cite{Teng2022} suggest; could help reshape estimations and tracking of debris within the ocean. There are currently methods for recreational and professional divers who complete the PADI Dive Against Debris course and log any data collection from scuba diving.

\subsection{Enhanced Techniques}

Enhanced techniques in this field are fundamental to addressing the issue of marine debris as it provides possible insights to how we can apply detection and tracking methods to real world scenarios.\\

\cite{Watanabe2019} proposed multiple sensors and data fusion techniques to revolutionise object detection in marine environments. Hyperspectral sensors were suggested by \cite{Maharjan2022} to identify plastics more clearly against other materials when the spectral properties alter from natural wear and degradation; they also suggest multispectral sensors to detect smaller debris. In the same year, \cite{Sánchez-Ferrer2022} suggested future research could explore other types of data such as sonar or lidar to heighten the detection and recognition of marine litter. \cite{Kylili2020} suggested the development of an autonomous image acquisition system that could collect footage of marine debris in real time and \cite{Watanabe2019} have also recommended using swarm control techniques for drones; underwater vehicles; cameras and satellite observations as a collective to monitor and detect ocean debris. They propose the use of reinforcement learning for optimally controlling multiple robots in this way.\\

Most of the research papers have not addressed their efforts to incorporate measures that protect the fragile underwater ecosystems as we attempt to use new methods for clean ups, as \cite{Fulton2019} mentions “accidental removal of biological entities such as plants or animals could damage the very ecosystems we hope to protect”. The authors are particularly concerned with environmental safety, as their research is based on the application of their model on robotics. They apply their model to a NVIDIA TM Jetson TX2 which is small enough for an AUV and produces minimal heat at a minimal electrical cost. Their team stated their intention to use their Aqua AUV \citep{Sattar2008} in the near future for real time object detection.

Despite this promising work, we feel that applying deep learning methods to automation is still an underexplored area. \cite{Hipolito2021} also briefly highlight the potential applications of testing and implementing such integration. \cite{Kasparavičiūtė2018} explore path planning solutions for plastic object retrieval using genetic programming (a method that allows the ability to learn for itself). They deployed this onto an AUV which is capable of diving underwater and used a range of sensors to detect plastic. They do however, register a multitude of challenges including consideration of exclusion zones; safety \& rollback; realistic path planning; battery consumption and general enhancement in the system. \cite{Fulton2019} also express the challenges of automation including “the development of navigation, exploration and manipulation strategies”. \\
However, upon conclusion of this review; we believe that in order to begin applying machine learning to automation for debris collection, researchers must first establish a much more in depth database and consider the many gaps in the research for detection and mapping algorithms.
\\

\cite{Acuña-Ruz2018} suggest using the results of their study on marine debris on beaches, where they combined three algorithms (RF, SVM \& LDA) to create an environmental management strategy for beach cleans and other means of decision making programs regarding beach debris. They particularly emphasise further developing AI methods and hyperspectral sensors for the classification of marine debris using satellite imagery. 
\\

Before testing any model in real world scenarios, it could be beneficial to create realistic simulations representing all classes with an introduction of new types of debris to monitor how the model would perform and behave.

\subsection{Deep Learning Methods}

The next steps in this research could be reviewing and comparing the strongest algorithms currently available (Faster R-CNN, YOLOv3, YOLOv4 \& YOLOv5) for object detection with marine debris. This would then have the potential to develop the best performing model into a hybrid option.

To avoid the natural bias that CNNs may produce; transformers could be explored further in combination with CNNs; to achieve a strong performance they would require a large database and powerful computer resources. There is positive research to suggest strong results with this combination by \cite{Bazi2021}, \cite{Carion2020}, \cite{Liu2021} \& \cite{Touvron2021} but minimal research of transformer CNN hybrids in marine debris. 
\\

\cite{Kikaki2022} suggested several directions when testing their benchmark dataset MARIDA. The authors proposed exploring training their dataset on more advanced machine learning models or investigating the transferability of the models on the dataset to other regions and sensors. The authors also highlighted the importance of developing methods for tracking and predicting marine debris; especially within ocean circulation models so that as researchers we can all better understand the way marine debris travels.\\
\cite{Sannigrahi2022} suggest separating the training data and utilising a multi object detecting model  to differentiate between organic floating debris and floating plastic. However, once again as mentioned above this would require a larger dataset reflecting more diversity, as well as broadening the regions where data is obtained.\\
\cite{Zhou2023} highlights the development of high performance and low size networks using dilated parallel modules (DPMs) as a promising direction - particularly in one-stage small object detection networks to transcend network performance.\\
\cite{Liu2023} suggest exploration of real-time deployment, particularly of their refined YOLOv5 model; including testing the models on different hardware applications for practical applicability in real-world scenarios. 
\\

It is also worth mentioning that a few researchers including \cite{Xue2021b}, \cite{Zhou2023} and \cite{Liu2023} recommend enhancing the performance of CNNs specifically for marine debris identification such as exploring different backbones, introducing additional attention mechanisms and incorporating other techniques to boost model performance.

\subsection{Evaluation Metrics} 

Evaluation metrics are used to assess the results of the model, which provides researchers with a better insight on how well the model is performing. \cite{Sánchez-Ferrer2022} touch briefly on how evaluating their model using the mean average precision may not be fully capturing the performance of the model, therefore, including a diverse array of evaluation metrics with each model trained could be beneficial. \cite{Watanabe2019} suggest evaluating the performance quantitatively for various deep network structures on different hardware options. \\

Based on the literature review and the results, deep learning for addressing marine debris has achieved great strides in the last five years, however, in order to make significant progress, there is a substantial journey ahead for researchers in this field, with an abundance of novelties to confront.

\section*{Acknowledgements}

Declaration of generative AI and AI-assisted technologies in the writing process
During the preparation of this work the author(s) used [ChatGPT-4] in order to [FIND EFFECTIVE SYNONYMS AND IMPROVE THE WRITING QUALITY]. After using this tool/service, the author(s) reviewed and edited the content as needed and take(s) full responsibility for the content of the publication.

\section*{Appendix A}

Google Scholar was searched for “deep learning marine debris” OR “deep learning marine macro debris”. In IEEE we searched for “deep learning marine debris”. 

We have checked references.
Papers excluded from this survey paper were either before 2018 or explored micro or nano debris detection.
It is worth also mentioning there is another database called JeDI “The Jellyfish Database Initiative” which could be easily confused with JEDI "JAMSTEC E-library of Deep-Sea Images". Some researchers in oceanscapes might use JeDI for tracking jellyfish populations. \citep{Condon}.

\section*{Supplementary} \textit{Table 1 Metadata comparison of all analysed papers from 'Section 3. Deep learning techniques' in marine debris identification.}


\bibliography{export}

\section{Author Statement}
\paragraph{Zoe Moorton} Conceptualization, Methodology, Formal analysis, Investigation, Resources, Software, Validation, Data Curation, Writing - Original Draft, Visualization
\paragraph{Dr. Zeyneb Kurt} Validation, Writing - Review and Editing, Supervision
\paragraph{Dr. Wai Lok Woo} Writing - Review and Editing, Supervision

\section{Acknowledgments}
Once again, I would like to thank the following who made this study happen by generously volunteering their data.

\renewcommand\labelitemi{\small$\bullet$} 
\begin{itemize} 
\itemsep=-1pt 		
\itemindent=-3pt 	
\item Maria Oskarsson
\item Dr. Melanie Bergmann, Alfred Wegener Institute Helmholtz Centre for Polar and Marine Research
\item Minna Zakheou, Viking Divers Ltd
\item Nic Emery, Fifth Point Dive.
\item Philip Robinson, Ay Nik Community Divers.
\item Thank you to the diving communities who helped contribute imagery.
\end{itemize}

I would also like to acknowledge JAMSTEC and Pexels for providing an open-source database to the public, from which I obtained much of my data.
Thank you to Ya'el Seid-Green, NOAA Marine Debris Program, who once again provided me with invaluable insight, and to Laura Anthony, Elizabeth Duncan and Heather Coleman, NOAA Federal for all their help.

\end{document}